# Designing AI-based Conversational Agent for Diabetes Care in a Multilingual Context

*Completed Research Paper*

**Thuy-Trinh Nguyen**
ReML Research Group
Sydney, Australia
Trinh.Nguyen@reml.ai

**Kellie Sim**
ReML Research Group
Singapore, Singapore
Kellie.Sim@reml.ai

**Anthony To Yiu Kuen**
University of Glasgow
Singapore, Singapore
2427248T@student.gla.ac.uk

**Ronald R. O'donnell**
Arizona State University
Phoenix, USA
Ronald.Odonnell@asu.edu

**Suan Tee Lim**
National University Health System
Singapore, Singapore
suan_tee_lim@nuhs.edu.sg

**Wenru Wang**
National University of Singapore
Singapore, Singapore
nurww@nus.edu.sg

**Hoang D. Nguyen**
University of Glasgow
Singapore, Singapore
Harry.Nguyen@glasgow.ac.uk

## Abstract

*Conversational agents (CAs) represent an emerging research field in health information systems, where there are great potentials in empowering patients with timely information and natural language interfaces. Nevertheless, there have been limited attempts in establishing prescriptive knowledge on designing CAs in the healthcare domain in general, and diabetes care specifically. In this paper, we conducted a Design Science Research project and proposed three design principles for designing health-related CAs that embark on artificial intelligence (AI) to address the limitations of existing solutions. Further, we instantiated the proposed design and developed AMANDA - an AI-based multilingual CA in diabetes care with state-of-the-art technologies for natural-sounding localised accent. We employed mean opinion scores and system usability scale to evaluate AMANDA's speech quality and usability, respectively. This paper provides practitioners with a blueprint for designing CAs in diabetes care with concrete design guidelines that can be extended into other healthcare domains.*

# Introduction

Globally, the prevalence of diabetes is rising rapidly. According to the World Health Organisation, diabetes was the seventh leading cause of death in 2016. Between 2000 and 2016, there was a 5% increase in premature mortality from diabetes (World Health Organization 2020). Despite the alarming consequences of diabetes, one in three diabetics either is unaware of their condition or has poor control of their health (Health Hub 2019).

Diabetes necessitates self-care through medication, vitals tracking and education. Through good self-management, people with diabetes can reduce the risk of health issues arising from diabetes and improve their quality of life. Adherence to these self-care routines, however, requires on-going motivation and commitment as diabetes can become a chronic medical condition that can be controlled but not cured (Mitzner et al. 2013). Therefore, diabetes care demands technological convenience and interventions to facilitate diabetic patients in self-management, especially during the global shortage of healthcare workers.

Conversational agents (CAs) have gained interest for their potential in self-care support for patients with medical conditions. CAs are programs designed to simulate conversations with human beings by processing natural language inputs and generating relative responses (Khan and Das 2018). Over the last two decades, a solid body of evidence has shown the potential of using CAs for health-related purposes, such as significant improvements in physical activity, fruit and vegetable consumption, and accessibility to online health information (Laranjo et al. 2018). To date, CAs have endorsed self-care for chronic diseases via condition diagnosis, coaching and data monitoring among other functionalities (Schachner et al. 2020). For diabetes care where self-care, knowledge and education are fundamental in maintaining patients' health in the long run, CAs can be of impactful help. Apart from typical weekly appointments with professionals, diabetic patients are essentially on their own over 90% of the time (Schatz 2016). Their self-care routine often involves medication and diet adherence, exercise, and vitals tracking, all of which need to be executed in a timely and consistent manner. Together, following this routine in the traditional approach can be frustrating as these activities require patients to have a sophisticated memory and sufficient medical knowledge. Given the development of increasingly powerful and connected devices, smartphone-based CAs are now widely used in diabetes self-care for educational tasks, including information retrieval and coaching (Morrow et al. 2020), which can not only lessen the hassles of the diabetes care routine but also provide users with personalised features that foster a sustainable lifestyle. Conversational artificial intelligence (AI), which often comprise powerful modules such as natural language understanding, language generation, has further enhanced the capabilities of CAs in health-related fields. However, there have been limited attempts on defining the prescriptive knowledge about designing CAs in the healthcare domain in general, and diabetes care specifically as knowledge is often scattered and lacks generalisation power. Hence, in this paper, we aim to answer the following question:

> *What are the fundamental design principles in designing a CA in the healthcare domain?*

To address this question, we conduct a Design Science Research (DSR) project that seeks a solution to a real-world problem via an iterative process (Hevner et al. 2004). First, we identify the problems from existing literature. Then, we articulate three design principles for designing healthcare CAs that can become a solid foundation for future applications. Next, we instantiate the design principles and develop AMANDA - a multilingual diabetes care CA with a natural-sounding Singaporean accent by implementing the state-of-the-art language models. Our proposed CA is evaluated in terms of audio quality and usability using mean opinion scores and system usability scale respectively.

# Background

### *Technological Interventions for Diabetes Care*

In recent years, technology has introduced a paradigm shift in diabetes care from rigid general decision making to data-driven and personalised recommendations. A fast-growing body technology, AI, is claimed to be the future of diabetes care in the way it transforms the prevention, diagnosis and treatment of diabetes. Generally, the use of AI in diabetes care can be categorised into automated retinal screening, clinical decision support, predictive population risk stratification and patient self-management tools (Ellahham 2020). Among these categories, self-management tools in the form of mobile applications and telehealth contribute to the routine empowerment of diabetic patients.

Most diabetes care applications have a CA as their central piece. While the application deals with most structured tasks including data tracking and monitoring, the CA acts as a coach to address patients' questions and guides them to utilise the functionalities of the system. Having a CA to attend to patients' questions and retrieve information in real-time allows users to ask repeated questions without feeling hesitant (Morrow et al. 2020). Through personalised conversations and their life-like characteristics, virtual agents can cultivate long-term engagement by building rapport with the users (Grzybowski et al. 2020).

*Natural Language Capabilities of Conversational AI*

Among the components of conversational AI, natural language capabilities are the most fundamental. The broad domain of Natural Language Processing (NLP) encompasses all software that interprets or produces a human language in either spoken or written form. Within NLP, two subdomains, namely Natural Language Understanding (NLU) and Natural Language Generation (NLG), exist. While NLU extracts the meaning of input data based on the context and decides on the user's intent and entities, NLG generates responses in written or spoken formats. There are two major challenges associated with a pipeline NLG architecture:

- **The generation gap:** Generation gap refers to the mismatch between strategic and tactical components which results in early decisions in the pipeline having unforeseen consequences further downstream. For instance, a generation system might determine a particular sentence ordering during the sentence planning stage, but this might turn out to be ambiguous once sentences have been realised, and orthography has been inserted (Meteer 1991).

- **Generation under constraints:** Being an instance of the generation gap, this problem occurs when the output of a system has to match certain requirements such as limited output length. Formalising constraints might be possible at the realisation stage by stipulating the length constraint in terms of the number of words or characters, but it is much harder in the earlier stages, where the representations are pre-linguistic and their mapping to the final text is potentially unpredictable (Gatt and Krahmer 2018).

Throughout the years of Text-to-Speech (TTS) development, there have been several techniques used to perform speech synthesis in NLG. A popular approach is the use of a concatenation speech synthesis system that joins waveforms from different parts of recorded speech stored in a large speech database to generate new utterances. Unit selection is used to effectively select the appropriate sub-word units from the database to be used for concatenation to synthesise an acceptable speech (Gatt and Krahmer 2018). As the speech database only contains a single speaker, however, changing properties like the accent of the speech would require new data.

WaveNet is a generative model based on the PixelCNN architecture (van den Oord et al. 2016) that can generate audio waveforms resembling a natural human voice by modelling waveforms using an autoregressive neural network trained with actual speech recordings. However, it requires a considerable amount of computational power and time to synthesis speech. The Parallel WaveNet architecture serves as an improvement to the WaveNet model. The use of Probability Density Distillation allows the WaveNet model to generate 20 times faster than real-time and retains its quality in a natural-sounding voice (van den Oord et al. 2017).

Tacotron 2 is introduced as an end-to-end TTS synthesis model capable of synthesising speech directly from text (Shen et al. 2018). It consists of a feature prediction network and a modified WaveNet vocoder. The text is first converted into character embeddings with an encoder, before being consumed by a decoder to predict a mel-spectrogram. A modified WaveNet model acting as the vocoder will then take in the mel-spectrograms and invert them into waveforms that resemble human-like speech. A mel-spectrogram is a representation of sound with the Short-Time Fourier Transform function. To improve the performance and accuracy of the model, an architecture is proposed to use an Encoder-Attention-Decoder network with location-sensitive attention.

## Design Research Project

To design and evaluate a CA for patients with diabetes in multilingual communities, we followed a 4-stage Design Science Research (DSR) framework of Kuechler and Vaishnavi (2008). In particular, we first studied problems of the context and proposed three design principles (DPs) for designing a conversational agent in the health care sector. Based on the proposed design, we developed a multilingual CA with a Singaporean accent. Subsequently, we evaluated the voice quality and usability of the proposed agent using mean opinion scores (MOS) and system usability scale (SUS).

*Awareness of Problem*

To investigate the underlying problems of existing CAs in diabetes, we first identified relevant literature using search combinations that include search terms for conversational agents (e.g. "conversational agent", "dialogue assistant", "dialogue system", "virtual agent" and "chatbot") and those for diabetes care (e.g. "diabetes", "type 2 diabetes"). Then, we searched for systematic reviews on CAs in healthcare and chronic conditions (Laranjo et al. 2018; Schachner et al. 2020), and consulted their reference lists. Subsequently, we found 9 CAs in diabetes care whose characteristics are shown in Table 1. Next, we will analyse three major limitations of the identified CAs.

**Table 1. Characteristics of conversational agents in diabetes care**

| Study | CA[a] | Purpose | Input | Output | Language |
|---|---|---|---|---|---|
| Friedman et al. 1997 | TLC | Telemonitoring Coaching | Audio (C[b]) | Audio | English |
| Black et al. 2005 | DI@L-log | Data collection Telemonitoring | Audio (C) | Audio | English |
| Huang et al. 2018 | Chatbot | Data collection Coaching | Text | Text | English Chinese |
| Bali et al. 2019 | Diabot | Disease diagnosis | Text (C) | Text | English |
| Stephens et al. 2019 | Tess | Data collection Coaching | Text (C) | Text | English |
| Baptista et al. 2020 | Laura ECA[c] | Coaching | Text Audio (C) | Text Audio | English |
| Balsa et al. 2020 | Vitória ECA | Data collection Coaching | Text (C) | Text Audio | English Portuguese |
| Anastasiadou et al. 2020 | EVA | Coaching | Text | Text | English Spanish Bulgarian |
| Rehman et al. 2020 | MIRA | Disease diagnosis | Audio | Text Audio | English |

[a]CA: Conversational agent.
[b]C: Constrained input.
[c]ECA: Embodied conversational agent.

**Limitation 1: Limited Technical Capabilities (L1)**

Diabetes care CAs are facing several technical challenges, including (1) the dominance of constrained input, and (2) unnatural-sounding vocal output. First, the majority of the identified CAs in our study only supports constrained input. In particular, most telephone-based (TLC, DI@L-log), text message (Tess) and embodied CAs (ECAs), such as Laura and Vitória, rely on constrained input. While CAs using telephone and text message as the medium only require basic interactions via phone keypad to perform telemonitoring and data collection tasks, ECAs are limited to constrained input due to the complex lip-syncing process between the speech sound and agents' movements. Second, limited research has focused on diabetes care CAs with spoken output. For CAs to generate natural spoken responses, it requires the transformation of TTS technology. Several CAs in diabetes care, however, have received comments for their "metallic voice" and unnatural responses (Balsa et al. 2020). These technical challenges can directly affect the user experience.

**Limitation 2: Lack of Fail-safe Features (L2)**

In general, most CAs in healthcare require user interaction to perform their functional requirements. In the face of ambiguous input, the agent might respond with irrelevant information. More seriously, miscommunication between CAs regarding health information may cause harm (Nadarzynski et al. 2019). While users might trust the CA to understand their queries and provide accurate information all the time, certain precautions must be put in place to minimise miscommunication. In most existing CAs in healthcare, however, patient safety is rarely accounted for. Unconstrained user input allows for more conversational flexibility but also comes with a higher risk for potential errors, such as mistakes in natural language understanding, response generation, or user interpretation of these responses (Laranjo et al. 2018). Also, helpful as CAs in diabetes care can be, limitations lie in their nature that needs to be refined by human experts; therefore, fail-safe features are needed.

**Limitation 3: Lack of Linguistic Personal Touch (L3)**

The lack of linguistic personal touch is another major issue in healthcare CAs. This issue is a two-tier problem. First, there have been limited attempts in extending the diversity of languages for CAs in diabetes care. The existing collection of non-English speaking CAs is limited with a few agents communicating in Chinese (Huang et al. 2018), Portuguese (Balsa et al. 2020), Spanish and Bulgarian (Anastasiadou et al. 2020). Among the identified CAs, only 3 out of 9 support multiple languages.

Further, within linguistic sensitivity, localised accent is rarely accounted for. A recent study reveals a novel phenomenon on the relationship between linguistic insecurity and linguistic ownership in post-colonial countries with Singapore as a leading example (Foo and Tan 2019). The study indicates that although English has been the official working language of Singapore and the medium of instruction in education for the past 30 years, Singaporean suffer from linguistic insecurity. In other words, despite Singaporean's ability to use English in both daily and professional context, they often consider themselves inadequate English speakers due to their localised accent. This unique trait of post-colonial nations, however, remains unaddressed in the existing CAs.

From a linguistic perspective, patients prefer communication in their mother tongue with higher treatment satisfaction (Hemberg and Sved 2019). On the other hand, patients who face language barriers have an increased risk of non-adherence to the treatment plan (Flores 2006). Thus, for diseases that demand consistent self-management like diabetes, it is reasonable for diabetes care applications and CAs to incorporate linguistic sensitivity to eliminate obvious linguistic obstacles.

*Suggestions*

To mitigate the stated limitations, we propose three design principles (DPs) for CAs in the healthcare domain. A DP is described as "a statement that prescribes what and how to build an artefact in order to achieve a predefined design goal" (Chandra et al. 2015). DPs are fundamental components of design research that allow essential communication, offer guidelines, and generalise prescriptive knowledge. We articulated the following DPs based on the framework of Chandra et al. (2015):

**DP1:** Provide the CA with sufficient natural language capabilities that have adequate naturalness, uniqueness, and clarity in order for users to communicate efficiently with the CA

**DP2:** Provide the CA with sufficient knowledge in subject-matter areas in order for users to be informed of safe and accurate information

**DP3:** Provide the CA with multilingual cultured interactions in order for users to be understood and feel personalised

Subsequently, we developed six design features (DFs) to extend the prescriptive knowledge of the proposed DPs. First, as mentioned in the literature review, the core natural language capabilities include NLU and NLG which are responsible for deriving the user's intents and generating responses respectively. Therefore, we articulated the following DFs to instantiate DP1:

**DF1:** The CA should understand the patient's intent and provide appropriate assistance.

**DF2:** The CA should respond in natural-sounding textual and/or auditory messages.

Second, user safety is a critical aspect of healthcare applications (Laranjo et al. 2018). Thus, the CA should have sufficient knowledge to address the user's questions accurately. However, health problems can be different for individuals and require professional consultation, which is out-of-scope for CAs. Hence, the CA should account for fail-safe features to deal with uncertainties besides a broad knowledge base. Also, being transparent with the users about its limitations enhances the trustworthiness of the agent (Rheu et al. 2020). To instantiate DP2, we articulated DF3 and DF4:

**DF3:** The CA should apply and facilitate relevant knowledge in the health care domain.

**DF4:** The CA should clarify ambiguous patient's input and/or forward uncertainties to human experts.

Finally, apart from the discussed benefit of having linguistic sensitivity and personalisation, such as overcoming obvious linguistic barriers, CAs that offer culturally familiar communication styles (e.g., language, accent and slang) are perceived as more trustworthy (Rheu et al. 2020). Thus, we formulated the following DFs as instantiations of DP3:

**DF5:** The CA should provide multilingual interfaces for patients in a seamless manner.

**DF6:** The CA should incorporate localisation in its natural language capabilities.

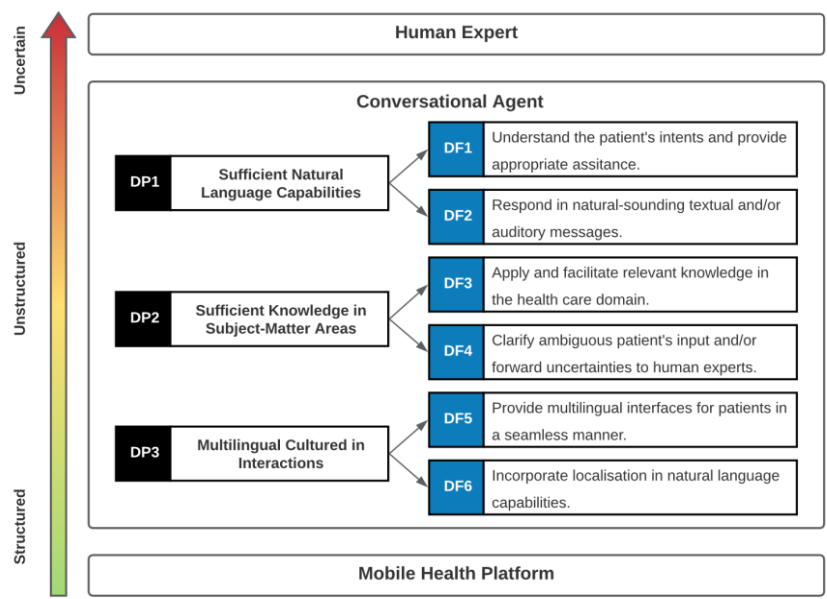

**Figure 1. Design Principles and Design Features**

In summary, Figure 1 demonstrates the three DPs and their corresponding DFs, and showcases the relationship between the CAs and other entities. In the bottom tier, mobile health platforms are recommended to handle structured tasks, including data tracking and setting reminders (Morrow et al.

2020). CAs, on the other hand, can provide medical advice to unstructured questions of the users. In the face of uncertainty, CAs should forward the patient's concerns to human professionals.

*Development*

In this section, we describe our instantiation of the design principles. We decided to focus on the Singaporean context because of (1) the prevalence of diabetes in Singapore, (2) the multicultural setting of the country, and (3) the special linguistic insecurity phenomenon as discussed in the previous section. In Singapore, about 440,000 residents who were 18 years and above had diabetes in 2014, and the number is estimated to reach 1,000,000 in 2050 (Health Hub 2019), which is largely attributed to the ageing population as the risk of diabetes increases with age. To best accommodate the Singaporean context, these CAs should be designed with a Singaporean accent due to their uniqueness. Singaporean English has been long recognised by linguists as the "standard" form of English in Singapore, where its syntax and lexicon are uniquely different from those of other English dialects.

This paper proposes AMANDA, a CA that answers queries on diabetes care, blood glucose monitoring and managing diabetes complications. With the name AMANDA, which stands for "Ask Me Anything on Diabetes Assistant", we create a virtual persona and identity to allow more personal feelings towards the agent. AMANDA supports conversations in English and Simplified Chinese (DF5) and features natural-sounding TTS capabilities to narrate replies and enhance the user interaction experience. Our Singaporean accent TTS is introduced to produce the English language spoken with the unique Singaporean accent (DF6). AMANDA is developed with health information approved by the Singapore National University Hospital (NUH).

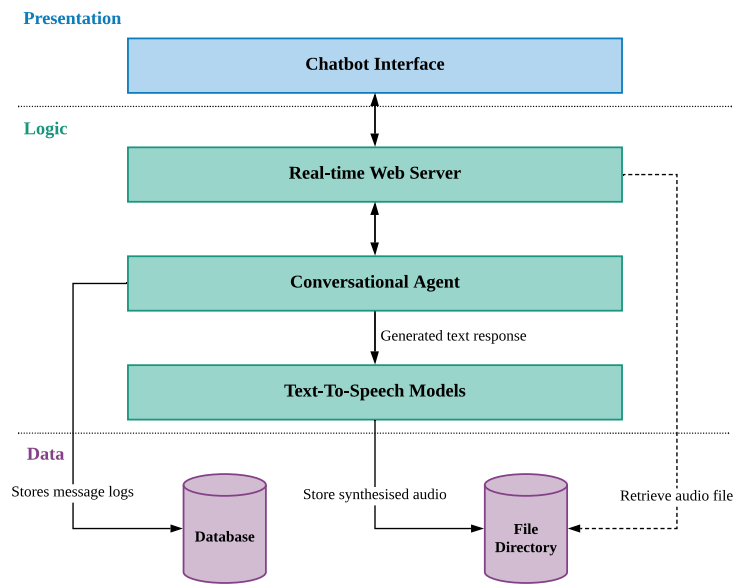

**Figure 2. A High-level Overview of the Conversational Agent Architecture**

*General Architecture*

The system follows a multi-tier architecture, consisting of a presentation, logic and data layer as shown in Figure 2. When the user sends a message through the interface, the message is transmitted to the logic layer for processing, where the CA interprets the message and produces a response to be sent back to the user interface. The data layer consists of a MongoDB storage system that stores the chat history between each user and the agent, as well as security logs generated from the various modules in the logic layer. The conversation history between the user and the agent is preserved locally on the user's device.

*Natural Language Capabilities*

Currently, AMANDA can process textual input, and generate both textual and auditory output. Therefore, our model employs the majority of components within conversational AI technologies, including NLU, dialogue management, NLG, and TTS. We implemented Rasa NLU to process and interpret the user's input (DF1). Rasa is currently among the most popular and trusted frameworks for machine learning-based NLU and dialogue management. For output generation, we employ a modified recurrent sequence-to-sequence network by conditioning WaveNet to generate time-domain waveforms, which is based on Tacotron 2 (Shen et al. 2018). Tacotron 2 is an integrated state-of-the-art end-to-end speech synthesis system that can generate speech of close-to-human quality in standard English. Synthesising natural-sounding speech of Singapore English, however, is strenuous from both data and modelling perspectives. The original encoder-decoder with location-sensitive attention mechanism of Tacotron 2 is prone to exposure bias. Its autoregressive sequence model appears backpropagating and intensifying speech imperfections of the data. Therefore, we adopted a bi-directional decoder regularisation to improve interactive updates between forward and backward decoders (Zheng et al. 2019). Figure 3 illustrates our TTS synthesis architecture.

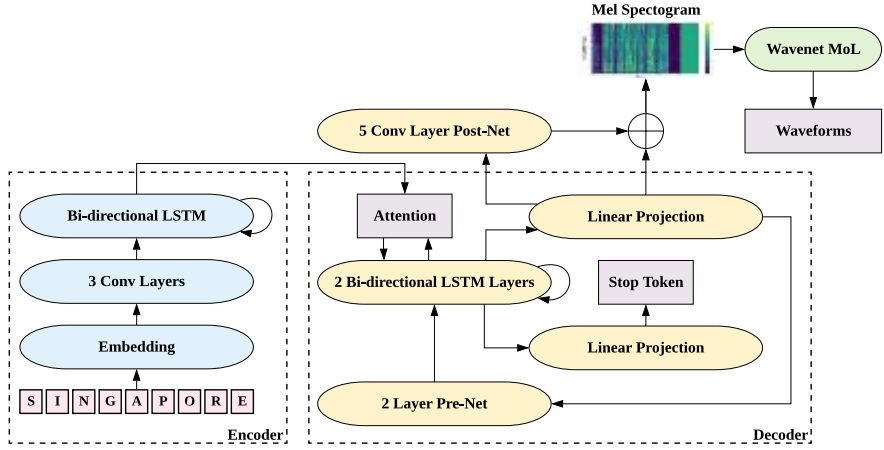

**Figure 3. TTS Synthesis Architecture for Localised Accent**

Our architecture consists of an encoder and a decoder illustrated in different boxes. The encoder converts a text sequence $x = \{x_1, x_2, ..., x_{T_x}\}$ into hidden representations $h = \{h_1, h_2, ..., h_{T_x}\}$. At each time step t, we compute the shared encoder as the following:

$$h_t = encoder(h_{t-1}, x_t)$$

The decoder is designed as bi-directional to predict the target mel-spectrograms $y = \{y_1, y_2, ..., y_{T_y}\}$. The hidden states $s_t$ are computed in bi-directional (both forward and backward) ways:

$$s_t = decoder(s_{t-1}, y_{t-1}, c_t)$$

where $c_t$ is a bi-directional context vector computed using a content-based attention mechanism to minimise exposure biases.

$$c_t = \sum_{k=1}^{T_x} \alpha_{tk} h_t$$

where the alignments $\alpha_{tk} = \frac{\exp(e_{tk})}{\sum_{i=1}^{T_x} \exp(e_{ti})}$ and the energy $e_{tk} = score(s_{t-1}, h_t)$.

In our model, we also introduce residuals by using Post-Net to improve overall output quality where $\widehat{y_r} = \mathfrak{F}(\hat{y})$ and $\widehat{y_p} = \hat{y} + \widehat{y_r}$.

We minimise the following loss for our optimisation problem:

$$\mathcal{L} = \mathcal{L}_{\overrightarrow{decoder}} + \mathcal{L}_{post-net} + \mathcal{L}_{\overleftarrow{decoder}} + \lambda \mathcal{L}_c$$

where $\mathcal{L}_{decoder} = \frac{1}{T_y}\sum_{t=1}^{T_y}(y_t - \hat{y}_t)$, $\mathcal{L}_{post-net} = \frac{1}{T_y}\sum_{t=1}^{T_y}\left(y_t - \widehat{y_{p_t}}\right)$, $\mathcal{L}_c = \frac{1}{T_y}\sum_{t=1}^{T_y}(\overrightarrow{s_t} - \overleftarrow{s_t})^2$ and the regularisation weight λ is set as 1.0.

Our proposed architecture provides state-of-the-art performance for the TTS synthesis of Singaporean English (DF2). Our Singaporean-accented TTS was trained from scratch using the Singapore English National Speech Corpus from the Infocomm Media Development Authority (IMDA), under the Singapore Open Data Licence. The dataset features the unique Singaporean accent from a female speaker, with a total length of approximately 8 hours. Table 2 shows an overview of the datasets used to train the TTS model.

**Table 2. Dataset for TTS training**

| Dataset | IMDA Corpus |
|---|---|
| No. of Sentences | 6,300 |
| Sampling Rate | 16 kHz |
| Total Length | 8 hours |
| Reference | Koh et al. 2019 |

The dataset is pre-processed by mapping the sentences in the transcript file to their respective audio file before they are loaded to the model for training. It is also shuffled and separated into two segments, where 90% of the dataset is used for training while the remaining 10% is used as test data. The training data is primarily used to train and fit the model, while the test data is used to assess its performance throughout the epochs.

To prepare for the training process, we employed a single NVIDIA T4 graphics card and CUDA 10.0 to support GPU-accelerated training. The TTS Synthesis was trained with the MozillaTTS project (Gölge 2020). We used Softmax as an attention norm function, batch size of 32, and the sampling rate at 16 kHz. The Adam optimiser was employed with an initial learning rate of $10^{-3}$ and decay after 5000 steps. Furthermore, the hyperparameters were fine-tuned after a series of experimental runs to seek the best results for the IMDA Speech Corpus as the dataset had not been explored before. The final training was concluded at 205,000 steps.

*Knowledge Base and Features*

To address DF3 regarding the knowledge base, the list of possible intents and medical information is provided by NUH, as well as some predefined questions for small talk. AMANDA answers queries on diabetes care, blood glucose monitoring and managing diabetes complications.

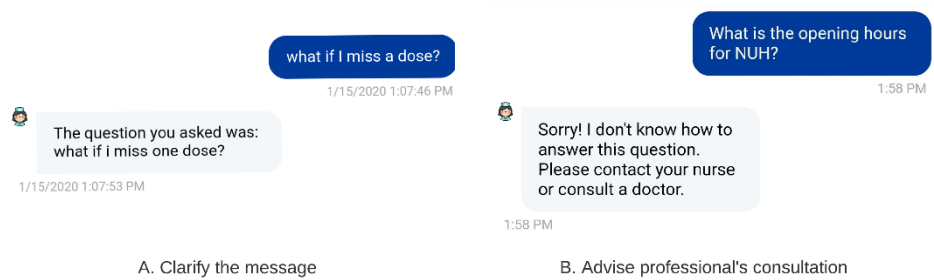

A. Clarify the message      B. Advise professional's consultation

**Figure 4. Fail-safe Features of AMANDA**

We introduced two fail-safe features for AMANDA, which are input clarification and expert advice extension (DF4). In the face of ambiguous input, AMANDA will verify the message before retrieving related information. AMANDA will respond with a confirmation message before answering a question, which will indicate the question predicted from the previous user message. If the question does not match what the user initially asked, they will have to rephrase their question for the agent to understand better. If the user asks a question beyond the expertise of the agent, AMANDA will prompt the user to consult their nurse or a doctor instead. Figure 4 demonstrates examples of the discussed fail-safe features.

Further, AMANDA also prompts the users with related questions that can be asked next, as shown in Figure 5, to enhance user experience.

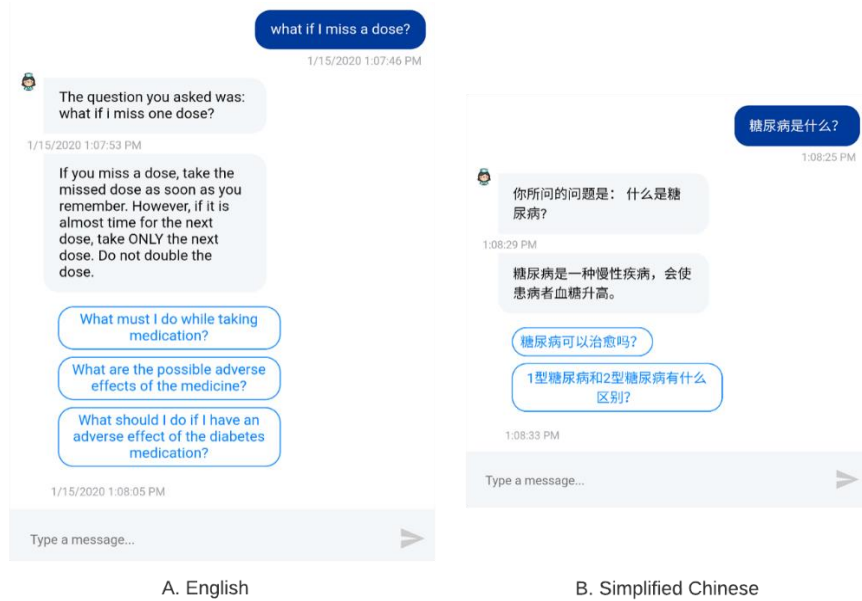

**Figure 5. Suggestion Feature of AMANDA**

*Evaluation*

To evaluate our CA, we conducted two types of evaluation on the audio quality using MOS, and the usability of AMANDA using SUS.

*Audio Quality Evaluation*

The synthesised audio outputs from the model closely resemble the ground truth under visual inspection, as shown in Figure 6. The task is comparably more challenging to train the model since the IMDA Corpus were smaller in size compared to other datasets and featured a unique accent different from the standard English phonemes.

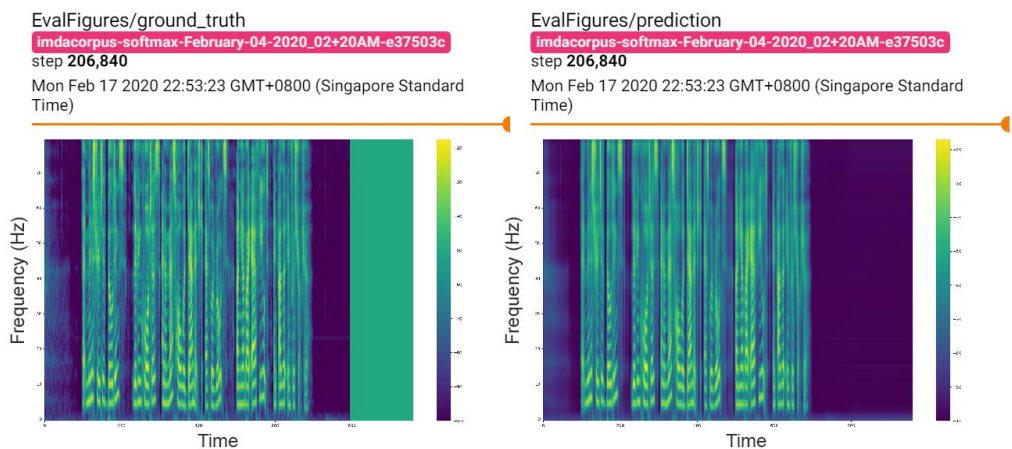

**Figure 6. Ground-truth and Predicted Mel-spectrograms Comparison on Singaporean-accented TTS**

To conduct the evaluation, a live audio recording from the IMDA Speech Corpus, which was used as training data for the model, was labelled as the real samples. We used our TTS to synthesis three types of audio samples: exact contents with the real samples, similar contents, and unseen contents.

There were 20 native judges recruited for our evaluation. They first listened to the real samples and paid attention to the speech characteristics of the speaker in the dataset. Next, the participants listened to the TTS generated samples and fill in a survey for each sample to evaluate the naturalness, Singaporean accent, and quality of the audio samples. We computed the MOS as shown in Table 3.

**Table 3. Mean Score Opinion Results**

| Measure | Exact | Similar | Unseen |
| --- | --- | --- | --- |
| Naturalness | 4.45 ± .74 | 4.3 ± .71 | 3.45 ± 1.2 |
| Singapore Accent | 4.05 ± 1.02 | 3.9 ± .94 | 3.65 ± 1.15 |
| Quality | 4.3 ± .9 | 4.15 ± .73 | 3.2 ± 1.07 |

Overall, we reported the MOS of **4.07** for naturalness, **3.98** for accent uniqueness, and **3.88** for clarity. The evaluation is effective in showing the current progress of the TTS. The participants provided useful feedback on how accurate and capable the TTS model was in generating speech that resembles the female Singaporean speaker in the dataset. More training should be done to minimise the gap in quality between seen and unseen utterances. The evaluation can also be improved by including audio recordings synthesised by other models with different neural architectures.

*System Usability Scale (SUS)*

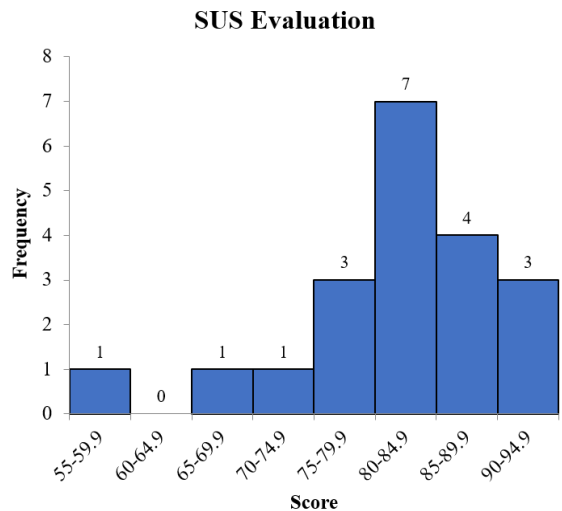

**Figure 7. SUS Scores Histogram on the Interface Usability Evaluation**

Besides, we use SUS, which is an industry-standard scale, to measure the perceived usability and learnability of a system. Originally developed by John Brooke (Brooke 1996), the SUS has become one of the popular choices to evaluate the product quantitatively. There were 20 nurses and clinicians invited to our experiment. The SUS score was computed to be **80.625**, which indicates that the interface is above the average score of 68 (Orfanou et al. 2015). Based on the histogram in Figure 7, 70% of the participants gave a score of 80 or higher. Several participants also praised the simple and practical layout design for the interface, which was easy to understand and use. The matching colours were appealing for the participants, and the interface looked modern and stylish.

Feedback suggested to include an indication of the usage of the buttons. A few participants voiced out that when AMANDA displayed relevant questions to be asked next, they did not understand what the bubbles were there for. A participant commented that the chat buttons should be located directly above the input message bar, similar to the ones in Facebook Messenger. Also, a text should be displayed (e.g. *"Here are more questions you might ask"*) to notify users that they can interact with the buttons to ask more questions. Currently, the narration for AMANDA's responses is being played right after the message appears. However, some participants were unable to comprehend the speech clearly for the first time and had to repeat the question to listen to the recordings again. Furthermore, some participants commented that there should be an option to disable the voice recordings, which gives the users more flexibility to choose whether they want the TTS functionality. The results demonstrate the adequate effectiveness of our TTS synthesis and CA for diabetes care.

## Discussion

In this paper, we conducted a DSR project for developing CAs in diabetes care. We proposed three design principles that address key limitations of existing literature. Subsequently, we instantiated six design features and developed AMANDA - a multilingual CA in diabetes care with state-of-the-art TTS synthesis for natural-sounding Singapore English. Our paper presents two main contributions being the prescriptive knowledge for designing CAs in the healthcare domain and a promising artefact with high usability.

The proposed design principles generalise the essential components of healthcare CAs to tackle major problems of existing CAs, which are (1) natural language limitations, (2) the need for a substantial knowledge base and fail-safe features, and the (3) lack of linguistic personal touch via multilingual interactions. Practitioners and developers can base their design on the prescriptive knowledge either to upgrade an existing CA or to develop future projects. Instantiations of the proposed DPs can be perceived as trustworthy because of their transparent fail-safe features and culturally familiar communication styles (Rheu et al. 2020). Our work draws attention to the scarce library of non-English and multilingual speaking CAs in healthcare in general, and in diabetes care specifically as the social impact of CAs should be considered thoroughly, from conceptualisation to real-world dissemination (Laranjo et al. 2018). Therefore, we urge future studies to consider linguistic and cultural sensitivity in their applications.

Further, we introduce a comprehensive conversational AI system with end-to-end NLU, NLG, and natural-sounding TTS for a localised accent. Our TTS architecture, which has gained promising feedback from the evaluation phase, can be a useful reference for future designs. We propose a modified architecture of the recurrent neural network decoder to predict a mel-spectrogram as an output from the text with a modified WaveNet vocoder to generate a natural-sounding waveform. Our work attempts to synthesise Singaporean accented TTS for English with the state-of-the-art end-to-end framework using the IMDA National Speech Corpus. With an overall MOS score of **4.07** for naturalness, **3.98** for accent uniqueness, and **3.88** for clarity, our design is an advancement in the diabetes care domain, where other agents received critics for their "metallic voice" and unnatural responses (Balsa et al. 2020). Further, a reported SUS score of **80.625** indicates that AMANDA is a highly usable diabetes assistant with natural-sounding TTS synthesis. Thus, future research on both healthcare CAs and natural language applications in Singaporean English can utilise our model.

## Conclusion

In this paper, we reported on our DSR project for designing CAs in the healthcare domain. We first identified three key problems of existing literature and proposed the corresponding design principles to address these limitations. Then, we instantiated the principles and developed AMANDA, which is a multilingual CA in diabetes care. We attempted to synthesis Singaporean-accented English TTS with the state-of-the-art end-to-end framework using the IMDA National Speech Corpus. Later, the instantiation was evaluated in terms of its audio quality using MOS and usability via SUS, which both reported satisfactory results. In summary, this paper has consolidated essential design features in healthcare CAs and proposed a comprehensive conversational AI system, which can serve as a useful reference for CA developers. Besides diabetes care, we endorse the use of CAs in other domains where self-management and personal touch are needed.

# References

Anastasiadou, M., Alexiadis, A., Polychronidou, E., Votis, K., and Tzovaras, D. 2020. "A Prototype Educational Virtual Assistant for Diabetes Management," in *2020 IEEE 20th International Conference on Bioinformatics and Bioengineering (BIBE)*, IEEE, pp. 999–1004.

Bali, M., Mohanty, S., Chatterjee, S., Sarma, M., and Puravankara, R. 2019. "Diabot: A Predictive Medical Chatbot Using Ensemble Learning," *International Journal of Recent Technology and Engineering (IJRTE)* (8:2), pp. 2277–3878.

Balsa, J., Félix, I., Cláudio, A. P., Carmo, M. B., Silva, I. C. E., Guerreiro, A., Guedes, M., Henriques, A., and Guerreiro, M. P. 2020. "Usability of an Intelligent Virtual Assistant for Promoting Behavior Change and Self-Care in Older People with Type 2 Diabetes," *Journal of Medical Systems* (44:7), p. 130.

Baptista, S., Wadley, G., Bird, D., Oldenburg, B., Speight, J., and My Diabetes Coach Research Group. 2020. "User Experiences With a Type 2 Diabetes Coaching App: Qualitative Study," *JMIR Diabetes* (5:3), p. e16692.

Black, L. A., McTear, M., Black, N., Harper, R., and Lemon, M. 2005. "Appraisal of a Conversational Artefact and Its Utility in Remote Patient Monitoring," in *18th IEEE Symposium on Computer-Based Medical Systems (CBMS'05)*, , June, pp. 506–508.

Brooke, J. 1996. *SUS-A Quick and Dirty Usability Scale (in "Usability Evaluation in Industry", PW Jordan, B Thomas, I McLelland, BA Weerdmeester (eds))*, London: Taylor and Francis, pp. 189–194.

Chandra, L., Seidel, S., and Gregor, S. 2015. "Prescriptive Knowledge in IS Research: Conceptualizing Design Principles in Terms of Materiality, Action, and Boundary Conditions," in *2015 48th Hawaii International Conference on System Sciences*, ieeexplore.ieee.org, January, pp. 4039–4048.

Ellahham, S. 2020. "Artificial Intelligence: The Future for Diabetes Care," *The American Journal of Medicine* (133:8), pp. 895–900.

Flores, G. 2006. "Language Barriers to Health Care in the United States," *The New England Journal of Medicine* (355:3), pp. 229–231.

Foo, A. L., and Tan, Y. 2019. "Linguistic Insecurity and the Linguistic Ownership of English among Singaporean Chinese," *World Englishes* (38:4), Wiley, pp. 606–629.

Friedman, R. H., Stollerman, J. E., Mahoney, D. M., and Rozenblyum, L. 1997. "The Virtual Visit: Using Telecommunications Technology to Take Care of Patients," *Journal of the American Medical Informatics Association: JAMIA* (4:6), pp. 413–425.

Gatt, A., and Krahmer, E. 2018. "Survey of the State of the Art in Natural Language Generation: Core Tasks, Applications and Evaluation," *arXiv [cs.CL]*. (http://arxiv.org/abs/1703.09902).

Gölge, E. 2020. "mozilla/TTS," *GitHub Repository*, GitHub. (https://github.com/mozilla/TTS).

Grzybowski, A., Brona, P., Lim, G., Ruamviboonsuk, P., Tan, G. S. W., Abramoff, M., and Ting, D. S. W. 2020. "Artificial Intelligence for Diabetic Retinopathy Screening: A Review," *Eye* (34:3), pp. 451–460.

Health Hub. 2019. "Diabetes in Singapore," *Health Hub*. (https://bit.ly/38wO9Uk).

Hemberg, J., and Sved, E. 2019. "The Significance of Communication and Care in One's Mother Tongue: Patients' Views," *Nordic Journal of Nursing Research*, SAGE Publications Sage UK: London, England, p. 2057158519877794.

Hevner, March, Park, and Ram. 2004. "Design Science in Information Systems Research," *The Mississippi Quarterly* (28:1), JSTOR, p. 75.

Huang, C.-Y., Yang, M.-C., Huang, C.-Y., Chen, Y.-J., Wu, M.-L., and Chen, K.-W. 2018. "A Chatbot-Supported Smart Wireless Interactive Healthcare System for Weight Control and Health Promotion," in *2018 IEEE International Conference on Industrial Engineering and Engineering Management (IEEM)*, IEEE, December. (https://doi.org/10.1109/ieem.2018.8607399).

Khan, R., and Das, A. 2018. "Build Better Chatbots," *A Complete Guide to Getting Started with Chatbots*, Springer. (https://link.springer.com/content/pdf/10.1007/978-1-4842-3111-1.pdf).

Koh, J. X., A. Mislan, K. Khoo, B. Ang, W. Ang, C. Ng, and Y. Y. Tan. 2019. "Building the Singapore English National Speech Corpus," in *2019 Proc. of INTERSPEECH*, , September, pp. 321-325.

Kuechler, B., and Vaishnavi, V. 2008. "On Theory Development in Design Science Research: Anatomy of a Research Project," *European Journal of Information Systems* (17:5), Taylor & Francis, pp. 489–504.

Laranjo, L., Dunn, A. G., Tong, H. L., Kocaballi, A. B., Chen, J., Bashir, R., Surian, D., Gallego, B., Magrabi, F., Lau, A. Y. S., and Coiera, E. 2018. "Conversational Agents in Healthcare: A Systematic Review," *Journal of the American Medical Informatics Association: JAMIA* (25:9), pp. 1248–1258.

Meteer, M. W. 1991. "Bridging the Generation Gap between Text Planning and Linguistic Realisation," *Computational Intelligence. An International Journal* (7:4), Wiley, pp. 296–304.

Mitzner, T. L., McBride, S. E., Barg-Walkow, L. H., and Rogers, W. A. 2013. "Self-Management of Wellness and Illness in an Aging Population," *Reviews of Human Factors and Ergonomics* (8:1), SAGE Publications, pp. 277–333.

Morrow, D. G., Lane, H. C., and Rogers, W. A. 2020. "A Framework for Design of Conversational Agents to Support Health Self-Care for Older Adults," *Human Factors*, p. 18720820964085.

Nadarzynski, T., Miles, O., Cowie, A., and Ridge, D. 2019. "Acceptability of Artificial Intelligence (AI)-Led Chatbot Services in Healthcare: A Mixed-Methods Study," *Digital Health* (5), p. 2055207619871808.

van den Oord, A., Kalchbrenner, N., Vinyals, O., Espeholt, L., Graves, A., and Kavukcuoglu, K. 2016. "Conditional Image Generation with PixelCNN Decoders," *arXiv [cs.CV]*, , June 16. (http://arxiv.org/abs/1606.05328).

van den Oord, A., Li, Y., Babuschkin, I., Simonyan, K., Vinyals, O., Kavukcuoglu, K., van den Driessche, G., Lockhart, E., Cobo, L. C., Stimberg, F., Casagrande, N., Grewe, D., Noury, S., Dieleman, S., Elsen, E., Kalchbrenner, N., Zen, H., Graves, A., King, H., Walters, T., Belov, D., and Hassabis, D. 2017. "Parallel WaveNet: Fast High-Fidelity Speech Synthesis," *arXiv [cs.LG]*, , November 28. (http://arxiv.org/abs/1711.10433).

Orfanou, K., Tselios, N., and Katsanos, C. 2015. "Perceived Usability Evaluation of Learning Management Systems: Empirical Evaluation of the System Usability Scale," *The International Review of Research in Open and Distributed Learning* (16:2), Athabasca University Press.

Rehman, U. U., Chang, D. J., Jung, Y., Akhtar, U., Razzaq, M. A., and Lee, S. 2020. "Medical Instructed Real-Time Assistant for Patient with Glaucoma and Diabetic Conditions," *NATO Advanced Science Institutes Series E: Applied Sciences* (10:7), Multidisciplinary Digital Publishing Institute, p. 2216.

Rheu, M., Shin, J. Y., Peng, W., and Huh-Yoo, J. 2020. "Systematic Review: Trust-Building Factors and Implications for Conversational Agent Design," *International Journal of Human--Computer Interaction*, Taylor & Francis, pp. 1–16.

Schachner, T., Keller, R., and V Wangenheim, F. 2020. "Artificial Intelligence-Based Conversational Agents for Chronic Conditions: Systematic Literature Review," *Journal of Medical Internet Research* (22:9), p. e20701.

Schatz, D. 2016. "2016 Presidential Address: Diabetes at 212° - Confronting the Invisible Disease," *Diabetes Care* (39:10), American Diabetes Association, pp. 1657-1663.

Shen, J., Pang, R., Weiss, R. J., Schuster, M., Jaitly, N., Yang, Z., Chen, Z., Zhang, Y., Wang, Y., Skerrv-Ryan, R., Saurous, R. A., Agiomvrgiannakis, Y., and Wu, Y. 2018. "Natural TTS Synthesis by Conditioning Wavenet on MEL Spectrogram Predictions," in *2018 IEEE International Conference on Acoustics, Speech and Signal Processing (ICASSP)*, , April, pp. 4779–4783.

Stephens, T. N., Joerin, A., Rauws, M., and Werk, L. N. 2019. "Feasibility of Pediatric Obesity and Prediabetes Treatment Support through Tess, the AI Behavioral Coaching Chatbot," *Translational Behavioral Medicine* (9:3), pp. 440–447.

World Health Organization. 2020. "Diabetes Fact Sheets," *World Health Organization*. (https://www.who.int/news-room/fact-sheets/detail/diabetes).

Zheng, Y., Wang, X., He, L., Pan, S., Soong, F. K., Wen, Z., and Tao, J. 2019. "Forward-Backward Decoding for Regularising End-to-End TTS," *arXiv [eess.AS]*, , July 18. (http://arxiv.org/abs/1907.09006).